\documentclass[runningheads]{llncs}

\usepackage[T1]{fontenc}

\usepackage{amssymb}
\usepackage{amsmath}
\usepackage{bbm}
\usepackage{booktabs}
\usepackage{pgfplots}
\pgfplotsset{compat=1.18}
\usepgfplotslibrary{fillbetween}
\definecolor{refBlue}{HTML}{0072B2}
\definecolor{refTeal}{HTML}{009E73}
\definecolor{gridGray}{HTML}{E0E0E0}
\usepackage{multirow}
 
\usepackage{MnSymbol}
\usepackage{pgfplots}
\pgfplotsset{compat=1.18}
\usepgfplotslibrary{fillbetween}
\usetikzlibrary{intersections}
\usepackage{graphicx,verbatim}
\usepackage[misc]{ifsym}
\usepackage{hyperref}
\usepackage{color}

\begin{document}

\title{DGRNet: Disagreement-Guided Refinement for Uncertainty-Aware Brain Tumor Segmentation}
\titlerunning{Disagreement-Guided Refinement for Uncertainty-Aware Segmentation}

\author{
Bahram Mohammadi\inst{1}\and
Yanqiu Wu\inst{1} \and
Vu Minh Hieu Phan\inst{2} \and
Sam White\inst{2} \and
Minh-Son To\inst{3} \and
Jian Yang\inst{1} \and
Michael Sheng\inst{1} \and
Yang Song\inst{4} \and
Yuankai Qi\inst{1} 
}
\authorrunning{B. Mohammadi et al.}
\institute{
Macquarie University, Sydney, NSW, Australia \\
\email{mohammadibahram71@gmail.com} \and
Adelaide University, Adelaide, SA, Australia \and
Flinders University, Adelaide, SA, Australia \and
University of New South Wales, Sydney, NSW, Australia
}
  
\maketitle

\begin{abstract}
\vspace{-2mm}
Accurate brain tumor segmentation from MRI scans is critical for diagnosis and treatment planning. Despite the strong performance of recent deep learning approaches, two fundamental limitations remain: (1) the lack of reliable uncertainty quantification in single-model predictions, which is essential for clinical deployment because the level of uncertainty may impact treatment decision-making, and (2) the under-utilization of rich information in radiology reports that can guide segmentation in ambiguous regions. In this paper, we propose the Disagreement-Guided Refinement Network (DGRNet), a novel framework that addresses both limitations through multi-view disagreement-based uncertainty estimation and text-conditioned refinement. DGRNet generates diverse predictions via four lightweight view-specific adapters attached to a shared encoder-decoder, enabling efficient uncertainty quantification within a single forward pass. Afterward, we build disagreement maps to identify regions of high segmentation uncertainty, which are then selectively refined according to clinical reports. Moreover, we introduce a diversity-preserving training strategy that combines pairwise similarity penalties and gradient isolation to prevent view collapse. The experimental results on the TextBraTS dataset show that DGRNet favorably improves state-of-the-art segmentation accuracy by $2.4\%$ and $11\%$ in main metrics Dice and HD95, respectively, while providing meaningful uncertainty estimates.

\end{abstract}

\section{Introduction}

Despite the significant improvements in brain tumor segmentation in recent years~\cite{rokuss2025voxtell,liu2025medsam3,shi2025textbrats,luo2025vividmed,xin2025text3dsam,xie2026tvpnet}, current methods face two key limitations. First, most approaches produce deterministic predictions without quantifying uncertainty. This is significant because acknowledging and, where possible, quantifying the level of uncertainty associated with AI models is highly relevant to treatment planning~\cite{wang2019aleatoric}. Although ensemble methods can provide an uncertainty score, they require training multiple models, which imposes high computational costs~\cite{abboud2024sparse}. Second, most existing methods~\cite{li2023lvit} apply text conditioning globally rather than focusing on specific uncertain regions where guidance is most valuable. 

To address these issues, we propose the Disagreement-Guided Refinement Network (DGRNet), a novel unified framework that achieves accurate segmentation and meaningful uncertainty quantification within a single model. The core idea is that prediction disagreement among diverse outputs serves as a powerful signal for identifying uncertain regions. DGRNet leverages disagreement to guide a targeted refinement process. Specifically, we introduce lightweight view-specific adapters that generate diverse segmentation masks from a shared backbone through feature-wise modulation. These adapters induce structured diversity at the semantic bottleneck level. The resulting multi-view predictions are aggregated through a disagreement module that fuses three complementary uncertainty metrics, namely prediction variance, pairwise disagreement, and predictive entropy, into a unified spatial uncertainty map that captures the multi-faceted nature of segmentation ambiguity.

The computed uncertainty map then drives a refinement module that selectively attends to unreliable regions. Through disagreement-aware spatial attention, features in high-uncertainty areas are amplified to enhance discriminability, while confident predictions remain stable. 
Also, we integrate clinical text guidance from radiology reports to resolve visual ambiguities that purely image-based methods cannot address. This text-conditioned modulation provides semantic context, such as tumor characteristics and location descriptors, that helps disambiguate challenging boundary regions. Moreover, to prevent view collapse, a critical failure mode in multi-view learning with shared representations, we introduce a diversity-preserving training strategy combining explicit bias initialization, pairwise similarity penalties, and gradient isolation. 
In contrast to conventional ensemble methods that require multiple models~\cite{supriyadi2025systematic} or Monte Carlo Dropout, which requires multiple stochastic passes, DGRNet produces calibrated uncertainty estimates in a single forward pass with only 5.8\% additional parameters over the baseline architecture, making it practical for real-time clinical deployment. Experimental results on the TextBraTS dataset show that DGRNet improves the state-of-the-art by 2.4\% and 11\% according to the main metrics Dice and HD95.
In summary, our main contributions  are below:

\begin{itemize}
    \item 
    We introduce an uncertainty-driven refinement approach that transforms prediction ambiguity from a passive diagnostic signal into an active mechanism for guiding targeted segmentation improvement in uncertain regions.

    \item 
    We propose a single-model framework that generates diverse predictions via lightweight view-specific adapters on a shared backbone, enabling ensemble-level uncertainty estimation without multiple models or stochastic passes.

    \item
    We develop an uncertainty-guided refinement module that selectively integrates radiological description text in ambiguous regions, and a diversity-preserving training strategy that prevents view collapse by bias initialization and similarity penalties. 
    
    \item 
    Extensive experiments on the TextBraTS benchmark demonstrate favorable improvements in segmentation accuracy over the state-of-the-art (SOTA) methods of 2.4\% and 11\% on the main metrics Dice and HD95, respectively. 
    alongside clinically meaningful uncertainty estimates.

\end{itemize}

\begin{figure*}[t!]
    \centering
    \includegraphics[width=0.99\linewidth]{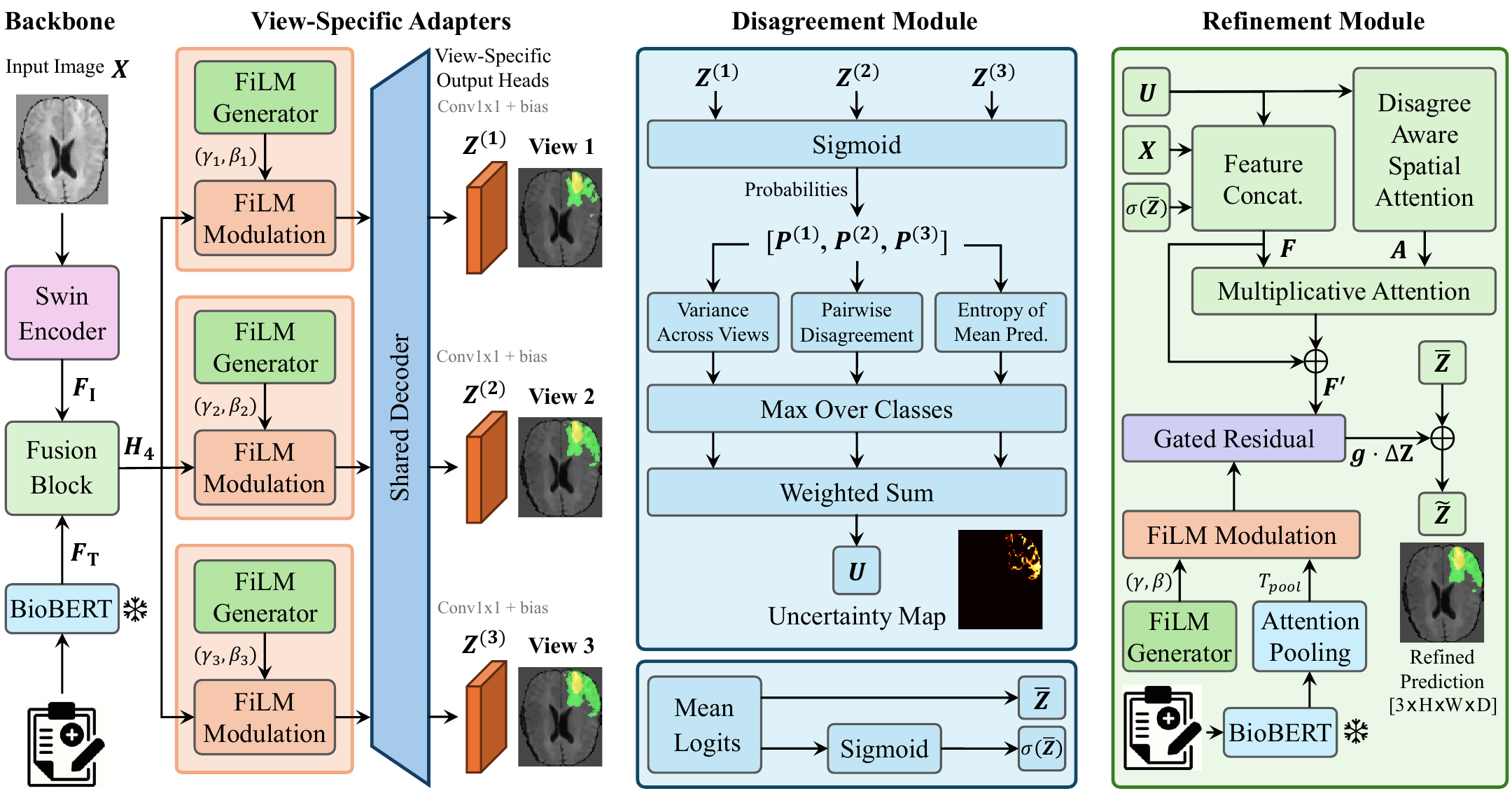}
    
    \vspace{-1mm}
       \caption{Overview of the DGRNet architecture. The model consists of four main components: (1) backbone extracts visual features via the Swin Transformer encoder, and textual features via the frozen BioBERT~\cite{lee_2020_biobert}, fused at the bottleneck, (2) view-specific adapters generate diverse predictions $Z^{(1)}$, $Z^{(2)}$, and $Z^{(3)}$ using learnable FiLM modulation with shared decoder and view-specific output heads (Sec.~\ref{sec::view_specific_adapters}), (3) disagreement module computes an uncertainty map $U$ from variance, pairwise disagreement, and entropy of mean predictions with learnable weights (Sec.~\ref{sec::disagreement_module}), and (4) refinement module uses disagreement-aware spatial attention and text-guided FiLM conditioning to produce a gated residual refinement $\Delta Z$, yielding the final prediction for all three sub-regions $\tilde{Z} = \hat{Z} + g \cdot \Delta Z$. (Sec.~\ref{sec::refinement_module})}
    \label{fig::outline}
\end{figure*}

\section{Method}
In this section, we first provide an overview of the proposed method, DGRNet, and the problem formulation. Then, we detail the main components.

\noindent\textbf{Overview.} 
Fig.~\ref{fig::outline} shows the main architecture of DGRNet, which consists of four collaborative components: (1) a shared encoder with text-guided feature extraction that provides rich multi-scale representations; (2) view-specific adapters that generate diverse predictions through learned specialization (Sec.~\ref{sec::view_specific_adapters}), (3) a disagreement module that quantifies prediction uncertainty through multiple complementary metrics (Sec.~\ref{sec::disagreement_module}), and (4) a text-guided refinement module that leverages both the disagreement map and semantic information to correct errors in high-uncertainty regions (Sec.~\ref{sec::refinement_module}). 

\noindent\textbf{Problem Formulation.}
Given a MRI volume $\mathbf{X} \in \mathbb{R}^{4 \times D \times H \times W}$ comprising four modalities (T1, T1ce, T2, FLAIR) along with an associated radiological text description $\mathbf{T}$, the goal is to predict the segmentation mask $\mathbf{Y} \in \{0,1\}^{C \times D \times H \times W}$ for $C=3$ hierarchical tumor sub-regions: the tumor core (TC), whole tumor (WT), and enhancing tumor (ET). Beyond segmentation, we seek to produce a voxel-wise uncertainty map $\mathbf{U} \in [0,1]^{D \times H \times W}$ that indicates prediction reliability at each spatial location.

\subsection{View-Specific Adapters}
\label{sec::view_specific_adapters}
To quantify prediction uncertainty without the computational overhead of full ensembles, we introduce a lightweight adaptation mechanism.

\noindent \textbf{View-Specific Adapters and Structured Diversification.} 
Our uncertainty estimation is based on meaningful diversity among views while leveraging shared representation learning. A significant challenge in shared encoder ensembles is view collapse, where multiple heads converge to identical predictions due to dominant shared gradient signals. To mitigate this, we propose a view-specific adapter mechanism that induces structured diversity via feature-wise linear modulation (FiLM)~\cite{dumoulin_2018_film} and decouples the output heads. Let $V$ denote the total number of views. For each view $v \in \{1, \dots, V\}$, we instantiate a learnable view embedding $\mathbf{e}_v \in \mathbb{R}^{d_b}$. This embedding serves as the input to a view-specific modulation generator, implemented as a multi-layer perceptron ($\text{MLP}_{\text{FiLM}}$), to generate affine scale $\boldsymbol{\gamma}_v$ and shift $\boldsymbol{\beta}_v$ parameters. These parameters are then used to modulate the global bottleneck feature map $\tilde{\mathbf{H}}_4^{(v)} = \mathbf{H}_4 \odot (1 + \boldsymbol{\gamma}_v) + \boldsymbol{\beta}_v$, where $\odot$ denotes element-wise multiplication. 

Besides the above modules, we also introduce two initialization strategies to further prevent view collapse:
% to diversify gradients. 
First, FiLM parameters are initialized with a small variance ($\sigma=0.1$) to start from a shared state. Second, we decouple the final projection layers $\mathbf{Z}^{(v)} = \text{Conv}_{1 \times 1}^{(v)}(\mathbf{D}_0^{(v)})$ and introduce a deterministic bias initialization: $\text{bias}^{(v)} = 0.02 \times (v - 1)$.

\vspace{-2mm}
\subsection{Disagreement Module}
\label{sec::disagreement_module}
This module acts as the aggregation component of DGRNet. The output of the previous module is used to generate two actionable outputs: a robust consensus segmentation and a comprehensive voxel-wise uncertainty map.

\noindent \textbf{Disagreement Module and Uncertainty Aggregation.} 
To robustly quantify prediction reliability, we aggregate the $V$ view-specific outputs into a unified uncertainty map $\mathbf{U}$ and a consensus segmentation. Specifically, we propose a learnable fusion of three complementary disagreement measures: (1) {Prediction variance} to capture the spread of predictions at the voxel level, effectively identifying regions where views split between confident positive and negative classes. (2) {Pairwise disagreement} to compute a soft Dice-based disagreement averaged over all unique view pairs for capturing structural and boundary inconsistencies that variance may mask. (3) {Predictive entropy} to compute the Shannon entropy~\cite{shannon_1948_mathematical} of the consensus prediction for scenarios where views agree but lack confidence.

\noindent \textbf{Learnable Uncertainty Fusion.} 
Rather than arbitrarily weighting these components, we employ a mechanism to learn the optimal combination for the specific anatomical targets. The final uncertainty map $\mathbf{U}$ is a weighted sum $\sum_{k \in \{\text{var}, \text{pair}, \text{ent}\}} w_k \cdot \mathbf{U}_k$, where $\mathbf{w} = \text{Softmax}(\boldsymbol{\theta}_w)$.

\noindent \textbf{Consensus Prediction.} 
For the final segmentation output, we compute the mean of the logits rather than the probabilities: $\bar{\mathbf{Z}} = \frac{1}{V} \sum_{v=1}^{V} \mathbf{Z}^{(v)}$. This approach, similar to geometric averaging in probability space, is preferred in ensemble theory for producing more accurate predictions.

\subsection{Refinement Module}
\label{sec::refinement_module}
Instead of treating uncertainty as a passive output, we actively leverage the uncertainty map $\mathbf{U}$ to spatially guide the refinement process. The core idea is that uncertain regions
correspond to visual ambiguities that require more focus and textual context.
This module operates as a residual correction block, refining the consensus prediction $\bar{\mathbf{Z}}$ into a final prediction $\hat{\mathbf{Z}}$.

\noindent \textbf{Disagreement-Aware Spatial Attention.} 
We first transform the uncertainty map into a spatial attention mask $\mathbf{A} \in [0, 1]^{D \times H \times W}$ to focus feature processing on unreliable regions. We then construct an informative input feature map $\mathbf{F}_{\text{in}} \in \mathbb{R}^{(4+C+1) \times D \times H \times W}$ by concatenating the raw image, current mean prediction, and uncertainty map $[\mathbf{X}; \sigma(\bar{\mathbf{Z}}); \mathbf{U}]$. 
This input is processed via a convolutional block to obtain $\mathbf{F}_1$, which is modulated by the disagreement attention $\mathbf{F}_1' = \mathbf{F}_1 \odot (1 + \mathbf{A})$. The term $(1+\mathbf{A})$ ensures a soft attention mechanism: high-confidence regions retain their original feature representation; however, uncertain regions are amplified to enhance feature discriminability.

\noindent \textbf{Text-Guided Feature Modulation.}
To resolve the visual ambiguities highlighted by $\mathbf{A}$, we inject semantic guidance from the clinical reports. We employ an attention-based pooling mechanism to extract relevant textual features $\mathbf{t}_{\text{pool}}$ that align with the current visual context. The pooled features serve as the conditioning signal for a FiLM layer, dynamically modulating the spatially attended visual features $\mathbf{F}_2 = \mathbf{F}_1' \odot (1 + \boldsymbol{\gamma}_t) + \boldsymbol{\beta}_t$

\noindent \textbf{Gated Residual Output.} 
The refined features $\mathbf{F}_2$ are processed via two additional convolutional blocks to produce a residual correction map $\Delta\mathbf{Z}$. We introduce a learnable scalar gate $g = \sigma(\theta_g)$ to control the magnitude of this correction $\Delta\mathbf{Z} = g \cdot \text{Conv}_{1 \times 1}(\mathbf{F}_4)$. Finally, the final prediction is obtained $\hat{\mathbf{Z}} = \bar{\mathbf{Z}} + \Delta\mathbf{Z}$.

\vspace{-2mm}
\subsection{Training Objective}
\vspace{-1mm}
The optimization of DGRNet balances three competing goals: accurate consensus segmentation, stable view diversity, and meaningful uncertainty estimation. The multi-objective loss function, which is composed of segmentation supervision and uncertainty-aware regularization terms, is formulated as:
\begin{equation}
    \begin{gathered}
        \mathcal{L}_{\text{total}} = \mathcal{L}_{\text{refined}} + \lambda_a \mathcal{L}_{\text{aux}} + \lambda_c \mathcal{L}_{\text{disagree}} + \lambda_v \mathcal{L}_{\text{diversity}}
    \end{gathered}
\end{equation}
This configuration establishes a self-regulating system, where $\mathcal{L}_{\text{refined}}$ and $\mathcal{L}_{\text{aux}}$ drive accuracy, $\mathcal{L}_{\text{disagree}}$ targets hard examples, and the $\mathcal{L}_{\text{diversity}}$ pair maintains the optimal variance window for uncertainty estimation.

\vspace{-2mm}
\section{Experiments}

\subsection{Experimental Settings}

\noindent \textbf{Implementation Details.}
We conduct the experiments on a single NVIDIA RTX A6000 GPU using PyTorch~\cite{paszke_2017_pytorch} with MONAI~\cite{cardoso_2022_monai}. We train for $200$ epochs with batch size of 1 using Sharpness-Aware Minimization (SAM)~\cite{foret_2021_sam} with SGD~\cite{nesterov_1983_sgd} as the base optimizer (lr=$0.1$, momentum=$0.9$).

\noindent \textbf{Dataset.}
We evaluate DGRNet on the TextBraTS dataset~\cite{shixia_2025_textbrats}, which is built based on the BraTS (Brain Tumor Segmentation) 2020  segmentation challenge training set~\cite{menze_2014_brats}. BraTS consists of multi-modal MRI scans with four co-registered sequences: T1, T1ce, T2, and FLAIR. Following standard protocols, we segment three hierarchically nested sub-regions: enhancing tumor (ET), tumor core (TC), and whole tumor (WT). 

\noindent \textbf{Evaluation Metrics.}
We report the Dice Similarity Coefficient~\cite{dice_1945_measures} and the 95th percentile Hausdorff Distance (HD95)~\cite{huttenlocher_2002_comparing} for segmentation. 
Alson, we assess the reliability of the generated uncertainty estimates $\mathbf{U}$ using: (1) Uncertainty Ratio, (2) Error Detection AUC, and (3) Sparsification Error (AUSE).

\begin{table*}[t!]
    \small
    \def\arraystretch{1}
    \setlength{\tabcolsep}{7pt}
    \caption{Comparison of DGRNet with the state-of-the-art methods on the TextBraTS dataset. $\dagger$ shows the results reproduced on the same platform as ours.
    }
    \vspace{-1mm}
    \centering
        \resizebox{\textwidth}{!}{
        \begin{tabular}{lcccccccc}
            \toprule
            
            \multicolumn{1}{c}{\multirow{2}{*}{Methods}} & \multicolumn{4}{c}{Dice (\%) $\uparrow$} & \multicolumn{4}{c}{HD95 $\downarrow$} \\
            \cmidrule(r{2pt}){2-5} \cmidrule(l{2pt}){6-9} &
            \multicolumn{1}{c}{ET} & \multicolumn{1}{c}{WT} & \multicolumn{1}{c}{TC} & \multicolumn{1}{c}{Avg.} &
            \multicolumn{1}{c}{ET} & \multicolumn{1}{c}{WT} & \multicolumn{1}{c}{TC} & \multicolumn{1}{c}{Avg.} \\

            \hline

            3D-UNet~\cite{olaf_2015_unet} & 80.4 & 87.3 & 81.6 & 83.1 & 6.11 & 10.51 & 8.93 & 8.17 \\
            
            nnU-Net~\cite{isensee_2021_nnu} & 82.2 & 87.5 & 82.6 & 84.1 & \underline{4.27} & 11.90 & 8.52 & 8.23 \\
            
            SegResNet~\cite{hsu_2021_brain} & 80.9 & 88.4 & 82.3 & 83.8 & 6.18 & 7.28 & 7.41 & 6.95 \\

            Swin UNETR~\cite{hatamizadeh_2021_swin} & 81.0 & 89.5 & 80.8 & 83.8 & 5.95  & 8.23 & 7.03 & 7.07 \\
            
            Nestedformer~\cite{xing_2022_nestedformer} & 82.6 & 89.5 & 80.2 & 84.1 & 5.08 & 10.51 & 8.93 & 8.17 \\

            TextBraTS~\cite{shixia_2025_textbrats} & 83.3 & 89.9 & 82.8 & 85.3 & 4.58 & \underline{5.48} & \underline{5.34} & \underline{5.13} \\

            \hline
        
            TextBraTS$^{\dagger}$ & 82.8 & 89.6 & 82.5 & 84.9 & 5.28 & 8.59 & 6.77 & 6.88 \\

            DGRNet (No Text) & \underline{84.6} & \underline{90.6} & \underline{84.6} & \underline{86.6} & 5.13 & 6.20 & 5.88 & 5.74 \\

            DGRNet (Full Mode) & \textbf{85.4} & \textbf{91.4} & \textbf{86.0} & \textbf{87.6} & \textbf{4.01} & \textbf{5.05} & \textbf{4.65} & \textbf{4.57} \\
        
            \bottomrule
            
        \end{tabular}}
        \vspace{-5mm}
    \label{tab::main_results}
\end{table*}

\vspace{-2mm}
\subsection{Comparison with SOTA methods} 
We compare DGRNet against the TextBraTS baseline, and other uni- and multi-modal SOTA methods. As shown in Table~\ref{tab::main_results}, DGRNet achieves a new SOTA performance, outperforming all comparison methods across all metrics. The alignment of high Dice scores with low HD95 values indicates that our model produces segmentation masks that are volumetrically accurate and topologically precise.

\noindent \textbf{Dice.}
DGRNet demonstrates a remarkable ability to segment complex tumor sub-regions. We achieve an average Dice score of $87.6\%$, outperforming the baseline by $2.4\%$. The most significant gains are observed in the most challenging classes, i.e., TC and ET, which are improved by $3.2\%$ and $2.1\%$, respectively. This validates the effectiveness of our method to mitigate ambiguous boundaries often found in the core and enhancing regions. Notably, DGRNet without text guidance already surpasses all existing methods.

\noindent \textbf{HD95.}
This metric provides strong evidence for the precision of our refinement approach. DGRNet achieves an average HD95 of $4.57$ mm, which is lower than TextBraTS by $0.56$ mm ($\approx11\%$ improvement). DGRNet maintains consistent, low-error boundaries across all sub-regions. This confirms that our uncertainty-guided refinement successfully corrects hard examples.

\subsection{Uncertainty Estimate}
A critical requirement for clinical adoption is that uncertainty must correlate with failure. The obtained results reveal a remarkable uncertainty ratio of $239.4\times$, indicating that the model is two orders of magnitude more uncertain about its errors ($\bar{u}_{\text{error}} \approx 0.018$) than its correct predictions ($\bar{u}_{\text{correct}} \approx 7.6 \times 10^{-5}$). The Error Detection AUC of $0.910$ confirms that the disagreement map acts as a highly effective binary classifier for segmentation errors. Moreover, the low Area Under Sparsification Error (AUSE = $0.0006$) shows that the uncertainty estimation capability of DGRNet is near-optimal for ranking pixel-wise reliability.

\begin{table}[t]
    \small
    \centering
    \setlength{\tabcolsep}{3pt}
    \caption{Ablation study of main components of our method.
    }
    \vspace{-1mm}
    \resizebox{\textwidth}{!}{
    \begin{tabular}{ccccc|cc}
        \toprule
        
        Multi-View & View Div. & Disagree Attn & Refinement & Text Cond. & Dice (\%) $\uparrow$ & HD95 $\downarrow$ \\
        
        \midrule
        $\times$ & $\times$ & $\times$ & $\times$ & $\times$ & 84.9 & 6.88 \\
        $\checkmark$ & $\times$ & $\times$ & $\times$ & $\times$ & 85.6 & 6.41 \\
        $\checkmark$ & $\checkmark$ & $\times$ & $\times$ & $\times$ & 86.2 & 5.94 \\
        $\checkmark$ & $\checkmark$ & $\checkmark$ & $\times$ & $\times$ & 86.6 & 5.48 \\
        $\checkmark$ & $\checkmark$ & $\checkmark$ & $\checkmark$ & $\times$ & 87.1 & 4.95 \\
        $\checkmark$ & $\checkmark$ & $\checkmark$ & $\checkmark$ & $\checkmark$ & \textbf{87.6} & \textbf{4.56} \\
        \bottomrule
    \end{tabular}}
    \label{tab::ablation_components}
\end{table}

\vspace{-2mm}
\subsection{Ablation Study}

\noindent \textbf{Effect of Components.}
To evaluate the contribution of each proposed component, we conduct an ablation study starting from a baseline. As shown in Table~\ref{tab::ablation_components}, we incrementally add: (1) multi-view prediction with text cross-attention, (2) view diversity through FiLM adapters, (3) disagreement-guided spatial attention, (4) the refinement module, and (5) text conditioning within refinement. Each component provides consistent improvements in both Dice score and HD95.

\begin{table}[t]
\small
\centering
\begin{minipage}{0.48\linewidth}
\centering
\caption{Ablation of numbers of views.
}
\vspace{-1mm}
  \begin{tabular}{cccccc}
    \toprule
    
    \multicolumn{1}{c}{\multirow{2}{*}{\#Views}} & \multicolumn{4}{c}{Dice (\%) $\uparrow$} \\

    & ET & WT & TC & Avg. \\
    
    \hline

    2 & 84.7 & 90.4 & 85.5 & 86.9 \\

    3 & 85.1 & 91.0 & \textbf{86.1} & 87.4 \\

    4 & \textbf{85.4} & \textbf{91.4} & 86.0 & \textbf{87.6} \\

    5 & 84.9 & 90.7 & 85.9 & 87.2 \\
 
    \bottomrule
  \end{tabular}
  \label{tab::ablation_view}
\end{minipage}
\hfill
\begin{minipage}{0.48\linewidth}
\centering
\caption{Ablation of diversity weight.
}
\vspace{-1mm}
  \begin{tabular}{cccccc}
    \toprule
    
    Diversity & \multicolumn{4}{c}{Dice (\%) $\uparrow$} \\
    
    Weights & ET & WT & TC & Avg. \\
    
    \hline

    0.0 & 83.5 & 90.0 & 84.2 & 85.8 \\

    0.5 & \textbf{85.4} & \textbf{91.4} & \textbf{ 86.0} & \textbf{87.6} \\

    1.0 & 85.3 & 90.2 & 85.4 & 87.0 \\

    5.0 & 84.0 & 90.1 & 85.1 & 86.4 \\
 
    \bottomrule
  \end{tabular}
  \label{tab::ablation_diversity}
\end{minipage}
\vspace{-4mm}
\end{table}

\noindent \textbf{Effect of Number of Views.}
Regarding Table~\ref{tab::ablation_view}, the performance improves as the number of views increases from 2 to 4, with the optimal performance achieved at 4 views. However, further increasing to 5 views leads to performance degradation. This pattern determines that too few views provide insufficient diversity for meaningful disagreement estimation, while too many views increase optimization difficulty and may disturb the disagreement signal

\noindent \textbf{Effect of Diversity Weight.}
As shown in Table~\ref{tab::ablation_diversity}, no diversity regularization causes view collapse, yielding the lowest performance. Moderate regularization achieves optimal results by maintaining meaningful view diversity. However, higher weights degrade performance by enforcing too much disagreement.

\subsection{Qualitative Analysis}
\begin{figure*}[t!]
    \begin{center}
    \includegraphics[width=0.75\linewidth]{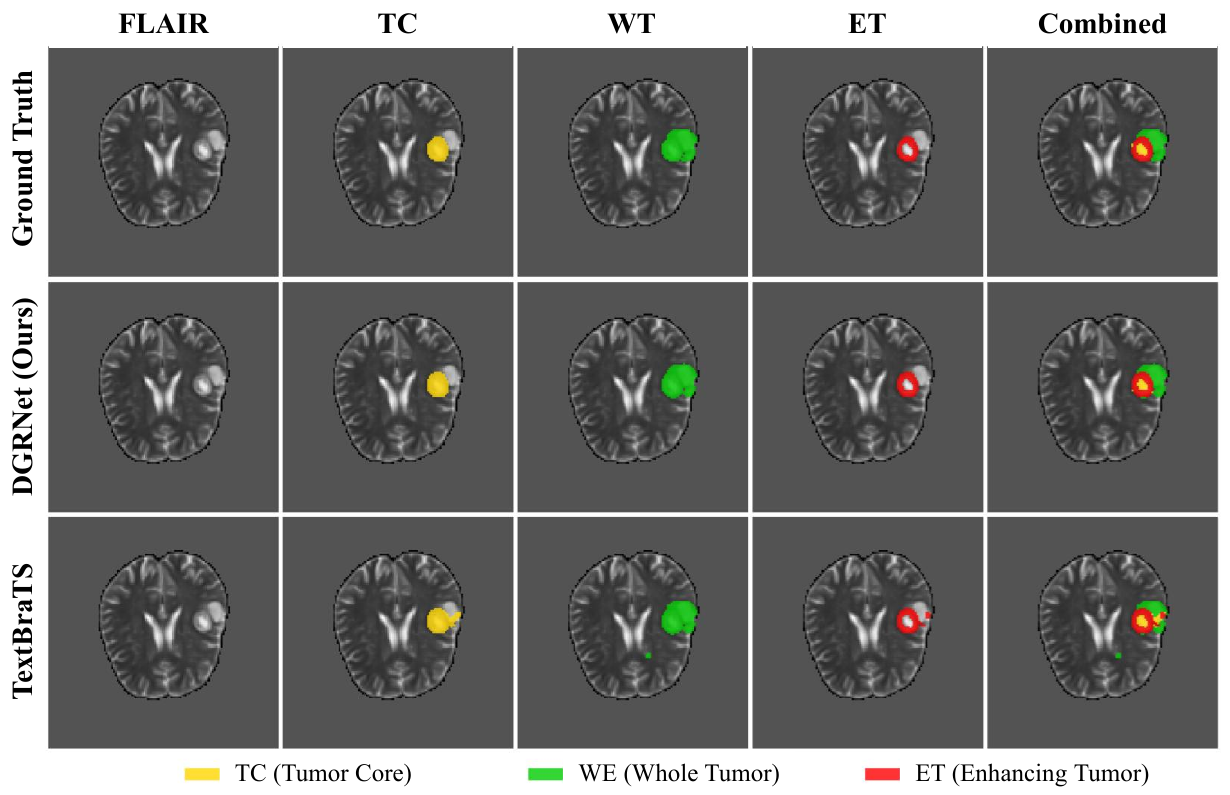}
    \end{center}
    \vspace{-7mm}
       \caption{Qualitative comparison with the baseline, using a representative sample that contains all of the sub-regions (TC, WT, and ET).}
    \label{fig::qualitative_comparison}
\end{figure*}

\begin{figure*}[t!]
    \begin{center}
    \includegraphics[width=0.75\linewidth]{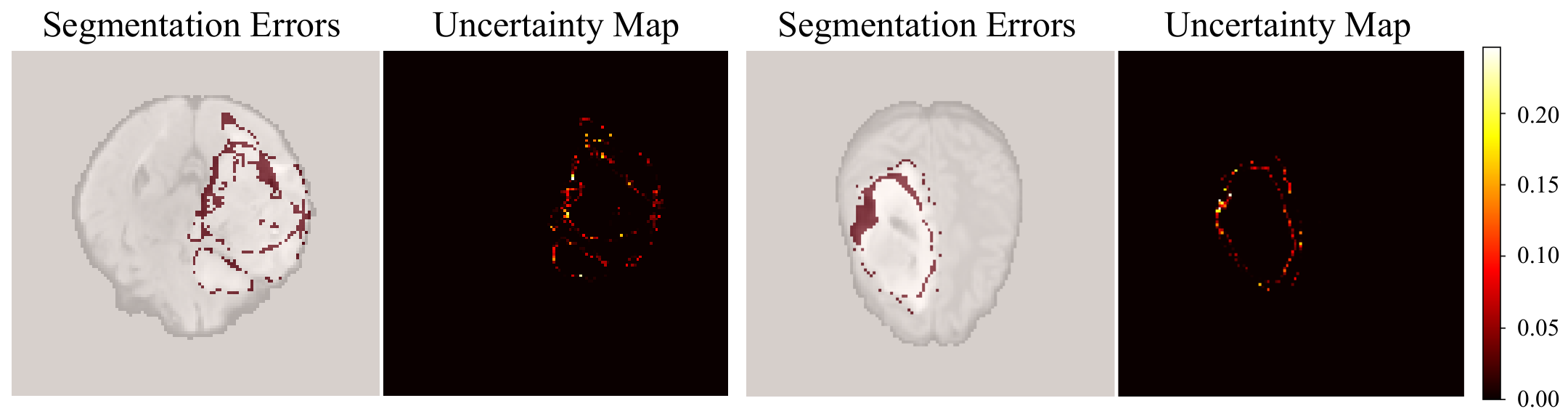}
    \end{center}
    \vspace{-7mm}
       \caption{The strong correlation between segmentation errors and the uncertainty map.}
       \vspace{-3mm}
    \label{fig::uncertainty_error}
\end{figure*}

Fig.~\ref{fig::qualitative_comparison} presents a qualitative comparison with the baseline model, showing the DGRNet's segmentation accuracy. Compared to TextBraTS, our method produces cleaner tumor boundaries with fewer false positives. Furthermore, Fig.~\ref{fig::uncertainty_error} demonstrates strong spatial correspondence between high-uncertainty regions and actual segmentation errors, with both concentrating along ambiguous tumor boundaries. This alignment confirms that DGRNet's uncertainty estimates are clinically meaningful for identifying regions where predictions may be unreliable and enable targeted review in clinical practice.

\vspace{-1mm}
\section{Conclusion}
We propose DGRNet, a framework that leverages prediction disagreement improve segmentation accuracy and also provides meaningful uncertainty estimates. By transforming uncertainty into an active refinement signal and integrating clinical text guidance in ambiguous regions, our method achieves state-of-the-art performance on the TextBraTS benchmark with $2.4\%$ Dice improvement and $11\%$ HD95 reduction, while producing uncertainty maps that correlate with actual errors. We believe this paradigm of uncertainty-guided refinement offers a promising direction for trustworthy medical image analysis.

\bibliographystyle{splncs04}
\bibliography{ref}

\end{document}